\documentclass[sigconf,nonacm]{acmart}

\AtBeginDocument{%
  \providecommand\BibTeX{{%
    \normalfont B\kern-0.5em{\scshape i\kern-0.25em b}\kern-0.8em\TeX}}}

\usepackage{url}
\usepackage{hyperref}
\usepackage[hyphenbreaks]{breakurl}

\begin{document}

\title{Reporting Eye-Tracking Data Quality: Towards a New Standard}


\author{Deborah N. Jakobi}
\authornote{Both authors contributed equally to this research.}
\email{jakobi@cl.uzh.ch}
\orcid{0000-0002-9719-6673}
\affiliation{%
  \institution{University of Zurich}
  \city{Zurich}
  \country{Switzerland}
}

\author{Daniel G. Krakowczyk}
\authornotemark[1]
\email{daniel.krakowczyk@uni-potsdam.de}
\orcid{0009-0009-5100-0733}
\affiliation{%
  \institution{University of Potsdam}
  \city{Potsdam}
  \country{Germany}
}

\author{Lena A. J\"ager}
\email{jaeger@cl.uzh.ch}
\orcid{0000-0001-9018-9713}
\affiliation{%
  \institution{University of Zurich}
  \city{Zurich}
  \country{Switzerland}\\
  \institution{University of Potsdam}
  \city{Potsdam}
  \country{Germany}
}


\begin{abstract}
Eye-tracking datasets are often shared in the format used by their creators for their original analyses, usually resulting in the exclusion of data considered irrelevant to the primary purpose. In order to increase re-usability of existing eye-tracking datasets for more diverse and initially not considered use cases, this work advocates a new approach of sharing eye-tracking data. Instead of publishing filtered and pre-processed datasets, the eye-tracking data at all pre-processing stages should be published together with data quality reports. In order to transparently report data quality and enable cross-dataset comparisons, we develop data quality reporting standards and metrics that can be automatically applied to a dataset, and integrate them into the open-source Python package \textit{pymovements} (\url{https://github.com/aeye-lab/pymovements}).
\end{abstract}



\maketitle

\label{intro}
\section{Introduction}
Eye-tracking-while-reading data has been used for a diverse range of purposes that range from addressing psycholinguistic research questions \cite{RAYNER2006613} to improving downstream applications in Natural Language Processing (NLP) \cite{barrett2018SequenceClassificationHuman, deng-etal-2023-pre} or biometric reader identification \cite{jager2020DeepEyedentificationBiometrica}. While data that is considered to be bad quality (e.g., low calibration scores) is often discarded in psycholinguistic research, more application-oriented research often requires vast amounts of data to train machine learning models and in some cases specifically requires data of non-optimal quality, as this can help to improve the model's generalizability to use case where only low-quality data is available \cite{prasse2020RelationshipEyeTracking}. Along the same lines, different use cases require data at different pre-processing stages. While some research questions typically focus on the analysis of aggregated reading measures on word- or sentence-level, other research uses sequence models to model scanpaths \cite{reich2022InferringNativeNonNative, vondermalsburg2011WhatScanpathSignature}, or models raw gaze data \cite{prasse2024ImprovingCognitivestateAnalysis}.
To completely enable the re-usability of datasets for use cases that were not initially considered by the creators, \textit{data at all pre-processing stages} needs to be made available. 
We propose a paradigmatic shift away from only publishing the filtered and pre-processed data that was used for the original data analysis towards a fully transparent and reproducible approach where no data is discarded. In particular, we propose to provide detailed metadata including data quality reports together with a fully transparent and reproducible pre-processing pipeline to enable the future users of the dataset to decide about their inclusion criteria and pre-processing. 

Eye-tracking data quality refers to the quality of the positional data collected during eye-tracking; that is how well the raw gaze data samples represent the actual movements of the eyes \cite{holmqvist2011EyeTrackingComprehensive}. Often, the quality is described in terms of spatial accuracy, temporal and spatial resolution (sampling frequency and precision), and data loss, all of which capture different aspects of the relationship between the true and the measured gaze position \cite{hessels2019EyeTrackingDevelopmental}. To date, there exist no concrete standards or metrics as to what is considered \textit{high quality} eye-tracking data for a specific use case. Many different and diverse factors are known to influence data quality such as the sampling frequency, the calibration procedures or the suitability of the stimulus size. Both, properties of the experimental setup that influence data quality and known data quality metrics such as calibration accuracy, need to be reported such that re-users can make informed decisions on whether the data might be useful for their research. 

The goal of our present research is to provide concrete suggestions and precise definitions of which properties of the setup and data quality measures to report, and we present a set of data quality metrics for dataset comparison. Following the FAIR principles of accessible, re-usable and interoperable data and code \cite{wilkinson2016FAIRGuidingPrinciples} and to ensure comparability across devices and datasets, reproducibility and transparency, we integrate our implementations into the Python package \textit{pymovements} \cite{krakowczyk2023PymovementsPythonPackage}.

\label{background}
\section{Related Work}
A first line of research on eye-tracking data quality investigates what factors influence data quality. For example, \citet{andersson2010SamplingFrequencyEyetracking} study how sampling frequency influences the detection of gaze events.
A second line of research actively seeks solutions to improve data quality either at recording time or post-hoc. Examples of this line of research are \citet{nystrom2013InfluenceCalibrationMethod}, who study how calibration procedures can be improved, or \citet{carr2022AlgorithmsAutomatedCorrection}, who investigate how vertical calibration drift can be corrected post-hoc using different algorithms \cite{Eyekit}. 
A third line of research seeks to standardize metadata reports about the experimental setup and procedure, and/or data quality. To date, there is no clear standard regarding which aspects of the setup/procedure of eye-tracking data collections should be reported, nor is there consensus on how to assess and report data quality. Oftentimes, information about the experimental setup and procedure that impact data quality, such as specifics on the calibration procedure, is missing. This issue of a lack of transparent data quality reporting was raised by different researchers \cite{dunn2023MinimalReportingGuideline, hessels2019EyeTrackingDevelopmental, nystrom2013InfluenceCalibrationMethod}.
A recent advancement in this area is the Python package \textit{pymovements}, that is in line with the open-source standards of complete transparency and reproducibility in combination with standardized documentation \cite{krakowczyk2023PymovementsPythonPackage}. The package allows users to select and apply different common eye-tracking pre-processing algorithms with freely chosen parameters but openly available source code which makes the pre-processing pipeline completely transparent and reproducible (if the parameters are shared by the users). In our work, we expand this package to include data quality reports that follow the same practices of transparency, reproducibility, open-source and standardized documentation.

\label{methods}
\section{Reporting Eye-tracking Data Quality}
We provide suggestions what metadata to report together with concrete implementations as well as a set of specific data quality metrics to create more transparent and re-usable eye-tracking datasets. Currently, we focus on trial- and session-level data quality reports for reading research conducted with different EyeLink devices but the general approach can be applied to any type of eye-tracking recordings and will be extended to other devices soon.

\textbf{Publishing data at all pre-processing stages.}
While traditional reading research analyses eye movement data almost exclusively as aggregated word-level reading measures (such as a word's first-pass fixation duration), more recent research started to use non-aggregated fixation data (e.g., scanpath analyses \cite{vondermalsburg2011WhatScanpathSignature, reich2022InferringNativeNonNative}) or even modelled the raw gaze samples \cite{jager2020DeepEyedentificationBiometrica, prasse2024ImprovingCognitivestateAnalysis}.
Eye-tracking research can continuously benefit if the data at all pre-processing stages including raw gaze samples is published. Most importantly, the raw gaze samples needs to be available in order to compute metrics such as the data loss ratio. When publishing eye-tracking data, it should be kept in mind that privacy issues can arise for both raw and aggregated data since eye gaze data at different preprocessing stages have been shown to exhibit strong idiosyncracies that can be used for biometric reader or viewer identification \cite{Holland_Komogortsev_2011, makowski2019DiscriminativeModelIdentifying, MakowskiTBIOM2021, MakowskiIJCB2020}.

\textbf{Reporting metadata.}
Each dataset should come with a set of standardized metadata on different levels (trial-, session-, and dataset-level, etc.). We therefore propose to automatically generate standardized metadata reports which we have integrated in \textit{pymovements}.
They include, for example, the following metadata: sampling rate, tracked eye(s), filter settings, date and time, total recording duration, eye tracker model and version information, and display resolution. 

\textbf{Reporting validation and calibration.}
Calibration and validation routines are included in the standard eye-tracking experiment procedure. Often, validations and/or re-calibrations are performed multiple times throughout experiments which should be reported as validation scores can point towards decreasing data quality. Our implementation includes information about the number and timestamps of calibrations/validations performed, the average and maximum scores, and the error label given by the recording device for each validation, and the tracked eye, as well as session-level settings such as the number of calibration points.

\textbf{Reporting data loss.}
Data loss refers to samples that were not recorded, for example, due to blinks or head movements. This means that the data contains less samples than there should be given the sampling frequency \cite{hessels2019EyeTrackingDevelopmental}. The reports we implemented include start and stop timestamps of blinks detected by the EyeLink software,
and the duration and number of samples for each blink, the overall data loss ratio including blinks and the data loss ratio caused by detected blinks. The difference between the two denotes samples lost due to unknown causes. A high blink ratio can further indicate that the participant felt uncomfortable.

\textbf{Data quality metrics.}
While the above mentioned reports are independent from the presented stimulus, we propose to additionally report \textbf{stimulus dependent metrics} that can be used to estimate the statistical validity of eye-tracking-while-reading data. In particular, we propose to report the word-skipping-rate, the background dwell-time, and the ratio of line jumps over multiple lines in standardized ways for all datasets. Furthermore, we propose to create metrics that incorporate well-studied psycholinguistic effects (e.g., the effect of age on the reading speed or the word length effect). 
If the same metrics are applied across datasets it will be possible to determine whether the reading behaviour recorded from a certain subject is indicative of inattentive reading, bad calibration or other factors. 

\label{future}
\section{Conclusion \& Future Work}
This work contributes to setting the foundation of a continued effort to i) increase the re-usability of eye-tracking data, ii) increase the transparency of eye-tracking data quality, and iii) enable comparisons between different eye-tracking datasets all of which strive towards making eye-tracking datasets align more with the FAIR principles \cite{wilkinson2016FAIRGuidingPrinciples}. We leave it to future research to extend the reporting standards and refine existing ones.

\label{impact}
\section{Broader Impact}
Reporting eye-tracking data quality can greatly improve any research in the field. While it is technically possible to obtain reports about, for example, blinks using the standard Data Viewer software to analyze EyeLink data files \cite{dataviewer}, these reports are eye-tracker-specific, and the exact implementation depends therefore on the manufacturer. In order to standardize these reports and shift towards a higher degree of transparency, it is of great value to use an open-source package that can be adapted to whichever device was used. Moreover, by facilitating the sharing of experiment details and gaze data at different preprocessing stages, we aim to pave the way for more research into ensuring high data quality and defining data quality standards across the field. As a consequence of these endeavours, it will be unavoidable to re-discuss and possibly change the role of manufacturers and proprietary restrictions on algorithms and data formats, and develop robust metadata and quality reporting standards which do not rely on closed-source proprietary software.

\begin{acks}
This work was partially funded by the Swiss National Science Foundation under grant 212276, by the German Federal Ministry of Education and Research under grant 01$\vert$S20043 and is based upon work from COST Action MultiplEYE, CA21131, supported by COST (European Cooperation in Science and Technology).
\end{acks}

\bibliographystyle{ACM-Reference-Format}
\bibliography{references}

\end{document}